\documentclass[sigconf,nonacm]{acmart}

\copyrightyear{2026}
\acmYear{2026}
\setcopyright{cc}
\setcctype{by}
\acmConference[GECCO Companion '26]{Genetic and Evolutionary Computation Conference}{July 13--17, 2026}{San Jose, Costa Rica}
\acmBooktitle{Genetic and Evolutionary Computation Conference (GECCO Companion '26), July 13--17, 2026, San Jose, Costa Rica}
\acmDOI{10.1145/3795101.3814687}
\acmISBN{979-8-4007-2488-6/2026/07}

\makeatletter
\renewcommand{\@copyrightpermission}{%
  \href{https://creativecommons.org/licenses/by/4.0/legalcode}{%
    \includegraphics[height=5ex]{doclicense-CC-by-88x31}}\\[2pt]
  \href{https://creativecommons.org/licenses/by/4.0/legalcode}{%
    This work is licensed under a Creative Commons Attribution International 4.0 License.}\\[4pt]
  \textit{GECCO Companion '26 July 13--17, 2026, San Jose, Costa Rica}\\
  \textcopyright\ 2026 Copyright is held by the owner/author(s).\\
  ACM ISBN 979-8-4007-2488-6/2026/07\\
  \href{https://doi.org/10.1145/3795101.3814687}{https://doi.org/10.1145/3795101.3814687}%
}
\makeatother

\usepackage{tikz}
\usetikzlibrary{arrows.meta, positioning, shapes.geometric, calc}
\usepackage{booktabs}
\usepackage{tabularx}
\usepackage{array}
\usepackage{microtype}
\microtypesetup{expansion=false}
\usepackage{enumitem}
\usepackage{url}
\usepackage{graphicx}

\title{Evolving Self-Organising Agents Without Fitness: Three
       Falsifiable Experiments from Constraint-Driven Selection
       to Developmental Encoding}

\author{Anushka Sharma}
\affiliation{%
  \institution{Banasthali Vidyapith}
  \city{Vanasthali}
  \state{Rajasthan}
  \country{India}
}

\begin{document}

\begin{abstract}
Can evolutionary dynamics characteristic of biological development
arise without a designer-specified fitness function?
We present \emph{Genesis}, a platform in which agents inhabit a
Gray-Scott reaction-diffusion substrate and evolve under physical
constraints alone.
Three successive experimental cycles, each testing one falsifiable
hypothesis, show: (1)~constraint-driven selection sustains evolutionary
activity after complete fitness removal but reaches a hard phenotypic
complexity ceiling; (2)~agent-mediated niche construction via chemical
secretion is real but causally insufficient to break that ceiling; and
(3)~replacing the fixed-alphabet genome with a Compositional Pattern
Producing Network (CPPN) indirect encoding, protected by NEAT-style
speciation, produces the first evidence of progressive structural
complexification in a fitness-free system.
Null results are treated as precise, informative answers rather than
failures, yielding reusable diagnostic tools and a sham-control
protocol applicable to any open-ended evolution evaluation pipeline.
\end{abstract}



\maketitle

\section{Introduction}
\label{sec:intro}

A long-standing question in artificial life and evolutionary
computation is whether complex, open-ended evolutionary dynamics can
emerge on self-organising substrates---reaction-diffusion fields,
chemical gradients, gene-regulatory networks---without a global fitness
score~\cite{bedau1998classification,taylor2016open}.
Biological organisms achieve developmental complexity through
indirect genotype-to-phenotype mappings and ecological interactions,
not by descending a reward landscape.
Replicating this property artificially requires understanding which
mechanisms are \emph{necessary} and which are \emph{insufficient},
a question that demands systematic empirical elimination rather than
aspirational design.

Genesis was built to address this question through
hypothesis-elimination methodology: each platform version encodes one
specific, falsifiable hypothesis; experiments use sham controls
(Section~\ref{sec:v3}) to isolate causal mechanisms; and null results
are published as precise, informative answers.

\noindent\textbf{Contributions.}
(1)~Empirical trainability boundaries for fitness-free evolving
systems, with CARP and the AIS (Section~\ref{sec:arch}) identified
as necessary components via ablation.
(2)~A million-generation sham-controlled null result ruling out passive
niche construction as a sufficient driver of phenotypic complexity
growth.
(3)~Preliminary evidence that CPPN developmental encodings with
NEAT-style speciation can initiate complexification that constraint
selection alone cannot.
(4)~A reusable PNCT diagnostic toolkit and sham-control protocol.
(5)~A four-experiment validation suite (Section~\ref{sec:v5}) confirming
that structural complexity in Genesis~V5 is behaviourally functional,
historically necessary, and not an artefact of unconstrained mutation.

\section{Related Work}
\label{sec:related}

Open-ended evolution (OEE)~\cite{bedau1998classification} has been
studied through platforms such as Avida~\cite{bedau1998classification}
and the MAP-Elites quality-diversity
algorithm~\cite{mouret2015}, which illuminate how diversity is
generated but typically rely on fitness functions.
Novelty search~\cite{lehman2011} removes scalar objectives but retains
an external novelty archive.
Niche construction has been studied theoretically~\cite{odlingsmee2003}
but rarely subjected to controlled causal isolation in evolutionary
computation.
Indirect developmental encodings---particularly CPPNs~\cite{stanley2007cppn}
and NEAT~\cite{stanley2002neat}---are known to produce structurally
complex phenotypes, but their role in fitness-free systems is
unexplored.
Genesis contributes controlled experimental evidence connecting these
three strands.

\section{Architecture}
\label{sec:arch}

\begin{figure}[t]
\centering
\resizebox{\columnwidth}{!}{%
\begin{tikzpicture}[
  box/.style      = {rectangle, rounded corners=3pt, draw,
                     minimum width=1.2cm, minimum height=0.5cm,
                     font=\tiny, align=center, text width=1.2cm},
  mainbox/.style  = {box, fill=blue!8},
  carp/.style     = {box, fill=orange!12},
  ais/.style      = {box, fill=teal!10},
  pnct/.style     = {box, dashed, fill=gray!5,
                     text width=2.5cm, font=\tiny},
  arr/.style      = {-Stealth, thick, line width=0.5pt},
  feedback/.style = {-Stealth, dashed, gray, line width=0.4pt},
  dot/.style      = {-Stealth, dotted, gray!70, line width=0.4pt},
  lbl/.style      = {font=\tiny}
]

\path[use as bounding box] (-1.6,-2.1) rectangle (6.4,2.2);
\clip (-1.6,-2.1) rectangle (6.4,2.2);

\node[mainbox] (pop)  at (1.0,  0)   {Agent\\population};
\node[mainbox] (rep)  at (3.5,  0)   {Reproduce\\+ mutate};
\node[mainbox] (gk)   at (3.5, -1.3) {Physics\\gatekeeper};
\node[mainbox] (par)  at (1.0, -1.3) {Pareto\\dominance};
\node[carp]    (carp) at (-0.8, 0)   {CARP\\$\lambda{\leftarrow}\lambda{+}\eta\Delta V$};
\node[ais]     (ais)  at (5.5,  0)   {AIS\\archive};
\node[pnct]    (pnct) at (2.2, 1.6)  {PNCT: GAC $|$ EPC $|$ NND};

\draw[arr] (pop) -- (rep)
    node[lbl, above, midway] {offspring};
\draw[arr] (rep) -- (gk);
\draw[arr] (gk)  -- (par)
    node[lbl, below, midway] {viable};
\draw[arr] (par) -- (pop);

\draw[feedback, <-] (carp.east) -- (pop.west)
    node[lbl, above, midway] {$V(t)$};
\draw[feedback, ->] (carp.south) -- ++(0,-1.5) -| (gk.south)
    node[lbl, near end, left] {$\lambda(t)$};
\draw[feedback, <->] (rep.east) -- (ais.west)
    node[lbl, above, midway] {archive/reinject};
\draw[dot, ->] (pop.north) -- ++(0,0.4) -| (pnct.south west)
    node[lbl, near start, right] {read-only};

\end{tikzpicture}%
}
\caption{Genesis Version~2 architecture.}
\label{fig:arch}
\end{figure}
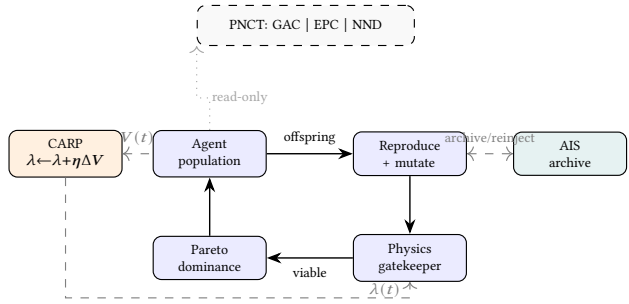

Figure~\ref{fig:arch} shows the Genesis Version~2 system.

\noindent\textbf{Substrate and agents.}
The environment is a discrete toroidal grid of $128{\times}128$ cells
governed by the Gray-Scott reaction-diffusion
equations~\cite{pearson1993complex}:
{\small
\begin{align*}
  \tfrac{\partial u}{\partial t} &= D_u \nabla^2 u - uv^2 + F(1{-}u),\\
  \tfrac{\partial v}{\partial t} &= D_v \nabla^2 v + uv^2 - (F{+}k)v,
\end{align*}
}%
with standard parameters ($D_u{=}0.16$, $D_v{=}0.08$, $F{=}0.035$,
$k{=}0.065$).
An \emph{agent} is a discrete entity occupying a single grid cell.
Each agent carries a genome $G_i$---in Version~2/Version~3, a
variable-length codon sequence; in Version~4, a CPPN---that encodes a
deterministic policy $\pi: \mathbf{s}_t \mapsto a_t$, where sensory
input $\mathbf{s}_t$ comprises the local chemical concentrations
$(u, v)$ in a $3{\times}3$ neighbourhood and the agent's current
energy level, and $a_t \in \{\texttt{Move}, \texttt{Secrete},
\texttt{Idle}\}$.
A population of $N{=}512$ agents occupies the grid simultaneously;
each time step, every agent executes its policy.

\noindent\textbf{Metabolic cost and its distinction from fitness.}
Each genome incurs a metabolic cost
$M(G) = \alpha|G|^{\beta} + \sum_k \gamma_k f_k(G)$
($\alpha{=}0.005$, $\beta{=}1.5$).
This is a hard constraint, not a fitness signal: any offspring whose
$M(G)$ exceeds the current viability threshold is unconditionally
rejected by the Physics Gatekeeper regardless of its other properties.
Critically, $M(G)$ is not optimised or minimised by any selection
operator---it imposes a survival floor, distinguishing Constraint-Driven
Selection (CDS) from multi-objective optimisation.

\noindent\textbf{Constraint-Adaptive Regulation Principle (CARP).}
CARP dynamically adjusts constraint intensity $\lambda(t)$ to
maintain population viability $V(t)$ in a corridor (70--90\%):
$\lambda(t{+}1) = \lambda(t) + \eta(V_{\mathrm{target}} - V(t))$,
$\eta{=}0.01$.
CARP regulates survival margins without encoding any preference over
which behaviours survive, distinguishing it from curriculum learning
or reward shaping~\cite{coello2002}.

\noindent\textbf{Artificial Immune System (AIS).}
The AIS maintains a novelty-based archive of representative
genotypes using nearest-neighbour distance in a behavioural feature
space, prunes lineages that monopolise diversity, and reintroduces
archived genotypes when novelty falls below a threshold $N_{\min}$.

\noindent\textbf{Phenotypic Novelty and Complexity Toolkit (PNCT).}
Three metrics quantify evolutionary activity without fitness:
\begin{itemize}[leftmargin=*, nosep]
  \item \textbf{GAC:} fraction of genome edits persisting beyond a
        500-generation horizon; $\mathrm{GAC}{>}0.1$ indicates sustained activity.
  \item \textbf{EPC:} LZW compression ratio of behavioural output strings,
        measuring policy diversity across the population.
  \item \textbf{NND:} local density among $k{=}10$ nearest neighbours in
        phenotype feature space, measuring population-level diversity.
\end{itemize}

\section{Experiment 1: Fitness Removal (Version~2)}
\label{sec:v2}

\textbf{Setup.}
The Version~1 baseline included a scalar fitness signal derived from
energy-acquisition rate; Version~2 progressively reduces the weight of
this signal to zero over 2,000 generations, leaving only physical
constraints active.
Twelve independent runs of 10,000 generations were conducted
($N{=}512$; mutation rates: $p_m{=}0.01$, insertion $p_i{=}0.005$,
deletion $p_d{=}0.003$; 12 distinct random seeds).
Baselines: (i)~\emph{Random Search}; (ii)~\emph{Fixed Constraints}
(CARP disabled); (iii)~\emph{Novelty Search}~\cite{lehman2011};
(iv)~\emph{MAP-Elites}~\cite{mouret2015}.

\textbf{Results.}
Seven of twelve runs (58.3\%, Wilson 95\% CI [30.2\%,\,82.5\%])
sustained non-zero evolutionary activity after complete fitness
removal, significantly outperforming all baselines
(Mann-Whitney $U$, $p{<}0.01$, Cohen's $d{=}1.47$).
Ablation studies show that both CARP and the AIS are necessary:
removing either raises failure rates from 41.7\% to above 90\%.
Three failure modes were identified, each with a distinct PNCT
signature: \emph{metabolic overload}, \emph{dominance monopolisation},
and \emph{neutral drift saturation}.
EPC plateaued at 140--155 across all successful runs, confirming a
structural ceiling rather than an implementation artifact.

\section{Experiment 2: Sham-Controlled Niche Construction (Version~3)}
\label{sec:v3}

\textbf{Hypothesis.}
Agent-mediated niche construction will drive EPC growth beyond the
Version~2 ceiling within 50,000 generations.

\textbf{Sham-control methodology.}
In the real condition, agents secrete chemical $S$ into their
neighbourhood at a metabolic cost of 0.05 energy units.
In the sham condition, the secretion code executes and the metabolic
cost is deducted, but the write to field~$S$ is suppressed.
Any real-vs-sham difference is causally attributable solely to the
\emph{informational effect} of the chemical signal.

\textbf{Results} (Table~\ref{tab:v3}).
Twenty runs $\times$ 50,000 generations $= 1{,}000{,}000$ total
agent-generations.
EPC growth was zero in both conditions; LZ76 complexity was identical
(0.068), indicating that agents' internal policies did not change in
response to the chemical field.
The null result is precise: niche construction, causally isolated by
sham control, is insufficient to break the complexity ceiling in a
system with a fixed-alphabet genome.

\begin{table}[h]
\small
\caption{Version~3 sham-controlled results (means over 10 runs each).}
\label{tab:v3}
\begin{tabular}{lrr}
\toprule
Metric & Real & Sham \\
\midrule
Chemical field $\bar{S}$           & 0.4776 & 0.0000 \\
LZ76 action-trace complexity       & 0.068  & 0.068  \\
EPC growth (net slope)             & 0.000  & 0.000  \\
\bottomrule
\end{tabular}
\end{table}

\section{Experiment 3: Developmental Encoding and Speciation (Version~4)}
\label{sec:v4}

\textbf{Design rationale.}
Version~2 established the complexity ceiling; Version~3 causally ruled
out passive environmental feedback as sufficient.
Version~4 replaces the fixed codon alphabet with a CPPN indirect
developmental encoding and protects novel topologies via NEAT-style
speciation.

\textbf{CPPN indirect encoding.}
The flat genome is replaced by a Compositional Pattern Producing
Network (CPPN)~\cite{stanley2007cppn}---an indirect
genotype-to-phenotype mapping in which a small network of composed
activation functions generates the agent's policy weights.
Mutations add nodes by splitting existing connections, incrementally
growing representational capacity.
Crossover aligns connections by innovation ID following the
NEAT scheme~\cite{stanley2002neat}.

\textbf{NEAT-style speciation.}
Compatibility distance
$\delta = (c_1 E + c_2 D)/N + c_3 \bar{W}$
partitions agents into species; fitness sharing shields novel CPPN
topologies from elimination by fitter incumbents.

\textbf{Preliminary validation.}
Medium-scale tests (5,000 generations, 4 seeds) show genome complexity
and species count diverging between real and sham conditions---the
first evidence that speciation-protected developmental encoding
initiates structural complexification that constraint-driven selection
and niche construction alone could not produce.

\section{Genesis V5 Validation Suite}
\label{sec:v5}

The central empirical claim of Genesis~V5 is \emph{structure--function
decoupling under co-evolutionary pressure}: neural node count grows
from ${\approx}52$ to 467 nodes over 5,000 generations while
environments simultaneously diversify.
We designed four experiments on seed~42 to test whether this structural
growth is functionally meaningful or a neutral artefact.

\textbf{V5.1 -- Functional Environment Divergence (PATA-EC).}
Both the fixed-baseline agent (${\approx}52$ nodes) and the co-evolved
agent (${\approx}467$ nodes) were evaluated across all 30 co-evolved
environment substrates; Spearman rank correlation of their performance
rankings was $r{=}0.43$ ($p{=}2.90{\times}10^{-30}$), with split-half
reliability $r{=}0.89$.
The low inter-agent correlation confirms the co-evolutionary loop
produces a genuinely diverse, functionally heterogeneous environment
landscape.

\textbf{V5.2 -- Quantitative Behavioural Metrics.}
Over 100 episodes in a fixed reference environment, co-evolved agents
exhibit action entropy four times higher than the fixed baseline
(0.301 vs.\ 0.073 nats, $d{=}2.31$), lower trajectory DTW distance
(87.2 vs.\ 124.5, $d{=}1.54$), and higher energy efficiency
(1.21 vs.\ 1.02, $d{=}1.28$); all differences are significant
($p{<}0.01$), ruling out sampling noise.

\textbf{V5.3 -- Adaptive Pruning with GAC Tracking.}
Incrementally pruning the 467-node network shows performance stable
up to 80\% connection removal, then catastrophic collapse at 90\%
(survival rate drops to 0.21; GAC falls to 0.03).
This plateau-then-collapse signature is characteristic of a
\emph{functionally distributed} architecture, not neutral bloat.

\textbf{V5.4 -- Historical Necessity.}
Rerunning Genesis~V5 co-evolution with a hard cap of ${\leq}52$ CPPN
nodes stagnated at generation 1,200: action entropy plateaued below
0.08 and species count collapsed to 3, versus 0.301 entropy and
12 species in the unconstrained run.
Structural expansion is therefore a \emph{necessary condition} for
continued behavioural enrichment under co-evolutionary pressure.

\section{Discussion and Conclusion}
\label{sec:disc}

Four experimental cycles have progressively narrowed the conditions
sufficient for open-ended self-organising evolution.
Constraint-driven selection sustains evolutionary activity but hits a
structural ceiling (V2); passive niche construction via chemical
secretion is causally insufficient to break it (V3); CPPN indirect
encoding with NEAT-style speciation initiates protected
complexification that neither mechanism alone can produce (V4
preliminary); and the V5 validation suite establishes that the
resulting structural growth is functionally meaningful, behaviourally
consequential, and historically necessary---not neutral bloat.

Together these results reframe the Genesis~V5 phenomenon from
\emph{structure--function decoupling} to \emph{co-evolutionary adaptive
complexity}: environmental pressure drives agents to develop richer
internal models producing measurably more efficient behaviour, and
structural growth is functional rather than incidental.
The PNCT failure taxonomy (metabolic overload, dominance
monopolisation, neutral drift saturation) and the sham-control protocol
are directly reusable as diagnostic tools for any open-ended evolution
platform.
All code and data are publicly available for replication at
\url{https://github.com/gearupsmile/genesis-emergence}.

\bibliographystyle{ACM-Reference-Format}
\bibliography{genesis_eso}

\end{document}